\documentclass[10pt,journal,compsoc]{IEEEtran}

\usepackage{cite}
\usepackage{amsmath,amssymb,amsfonts,bm}
\usepackage{mathtools}
\usepackage{graphicx}
\usepackage[font=footnotesize]{caption}
\usepackage[table]{xcolor}
\usepackage{balance} 
\usepackage{csvsimple}
\usepackage{multirow}
\usepackage{array}
\usepackage{adjustbox}
\renewcommand{\arraystretch}{1.3}

\usepackage{url}
\usepackage[inline]{enumitem}
\usepackage{makecell}
\usepackage{subfig}
\usepackage{pifont}
\newcommand{\cmark}{\ding{51}}%
\usepackage[para]{threeparttable}
\usepackage{todonotes}

\newcommand{\textapprox}{\raisebox{0.5ex}{\texttildelow}}

\makeatletter
\newcommand*\titleheader[1]{\gdef\@titleheader{#1}}
\AtBeginDocument{%
  \let\st@red@title\@title
  \def\@title{%
    \bgroup\normalfont\footnotesize\centering\@titleheader\par\egroup
    \vskip1ex\st@red@title}
}
\makeatother

\title{Hardware-Aware DNN Compression via Diverse Pruning and Mixed-Precision Quantization}
\titleheader{\vspace{-55pt}This article is accepted for publication in IEEE Transactions on Emerging Topics in Computing (DOI: 10.1109/TETC.2023.3346944).\\This is the author's version which has not been fully edited and content may change prior to final publication.}%

\begin{document}
\bstctlcite{IEEEexample:BSTcontrol}
\author{
Konstantinos~Balaskas,
Andreas~Karatzas,
Christos~Sad,
Kostas~Siozios, \IEEEmembership{Senior Member, IEEE},\\
Iraklis~Anagnostopoulos, \IEEEmembership{Member, IEEE},
Georgios~Zervakis,
J\"org~Henkel, \IEEEmembership{Fellow, IEEE}
\IEEEcompsocitemizethanks{\IEEEcompsocthanksitem K. Balaskas, C. Sad and K. Siozios are with the Department of Physics, Aristotle University of Thessaloniki, Thessaloniki 54124, Greece. K. Balaskas is also with the Chair for Embedded Systems at Karlsruhe Institute of Technology, Karlsruhe 76131, Germany.
\IEEEcompsocthanksitem A. Karatzas and I. Anagnostopoulos are with the School of Electrical, Computer and Biomedical Engineering, Southern Illinois University Carbondale, Carbondale, IL 62901 USA.
\IEEEcompsocthanksitem G. Zervakis is with the Dept. of Computer Engineering \& Informatics, University of Patras, Patras 26504, Greece. This research was done when he was with the Karlsruhe Institute of Technology.
\IEEEcompsocthanksitem J. Henkel are with the Chair for Embedded Systems, Karlsruhe Institute of Technology, Karlsruhe 76131, Germany.}%
\thanks{Corresponding author: Konstantinos Balaskas (balaskas@kit.edu).}%
\thanks{This work is supported in parts by grant NSF CCF 2324854, by the German Research Foundation (DFG) project ``ACCROSS''  HE 2343/16-1 under the grant 428566201 and by the E.C. Funded Program ``SERRANO'' under H2020 Grant 101017168.}%
\thanks{Manuscript received October 13, 2022, revised July 31, 2023 and November 16, 2023}%
}

\markboth{IEEE Transactions on Emerging Topics in Computing}%
{K. Balaskas \MakeLowercase{\textit{et al.}}: Hardware-Aware DNN Compression via Diverse Pruning and Mixed-Precision Quantization}

\IEEEtitleabstractindextext{%

\begin{abstract}
Deep Neural Networks (DNNs) have shown significant advantages in a wide variety of domains.
However, DNNs are becoming computationally intensive and energy hungry at an exponential pace, while at the same time, there is a vast demand for running sophisticated DNN-based services on resource constrained embedded devices.
In this paper, we target energy-efficient inference on embedded DNN accelerators.
To that end, we propose an automated framework to compress DNNs in a hardware-aware manner by jointly employing pruning and quantization.
We explore, for the first time, per-layer fine- and coarse-grained pruning, in the same DNN architecture, in addition to low bit-width mixed-precision quantization for weights and activations.
Reinforcement Learning (RL) is used to explore the associated design space and identify the pruning-quantization configuration so that the energy consumption is minimized whilst the prediction accuracy loss is retained at acceptable levels.
Using our novel composite RL agent we are able to extract energy-efficient solutions without requiring retraining and/or fine tuning.
Our extensive experimental evaluation over widely used DNNs and the CIFAR-10/100 and ImageNet datasets demonstrates that our framework achieves $39\%$ average energy reduction for $1.7\%$ average accuracy loss and outperforms significantly the state-of-the-art approaches.
\end{abstract}

}
\maketitle
\IEEEpeerreviewmaketitle

\section{Introduction}\label{sec:intro}
During the last decade, Deep Neural Networks (DNNs) have been established as the
driving force in a wide range of application domains, such as object detection, speech recognition, virtual/augmented reality and more~\cite{Jouppi:ISCA:2017:datacenter}.
Modern DNN architectures are becoming increasingly more complex and deeper, with an ever increasing number of trainable parameters, striving to maximize the accuracy of modern applications.
In order to keep up with the increased computational complexity, hardware DNN accelerators~\cite{Chen:JSSC:2016:eyeriss} form an integral part of computing systems.
Such accelerators integrate thousands of multiply-accumulate (MAC) units, responsible for executing the required arithmetic operations~\cite{Amrouch:TCAD:2021}.
Even though this large number of MAC units, operating in parallel, results in improved performance, it comes with the burden of vast energy consumption, thus greatly affecting their integration in energy-constrained embedded systems~\cite{Amrouch:TCAD:2021}.

In order to balance the trade-off between accuracy and energy consumption, several DNN compression techniques have been exploited.
Noticeably, pruning and quantization have emerged as the most popular, mainly due to their effectiveness and simplicity.
Pruning consists of intelligently sparsifying a dense DNN, by either removing connections (i.e., fine-grained pruning~\cite{Han:NIPS:2015:weights,Lee:ICLR:2019:snip, Guo:arxiv:2016:dynamic}) or regular structures (i.e., coarse-grained pruning~\cite{Li:arxiv:2016:convnets,Gao:ICLR:2018:channel, Ma:TNNLS:2021:non}).
Quantization~\cite{Choi:arxiv:2018:pact, Jacob:CVPR:2018:quantization} on the other hand, lowers the arithmetic precision of operands (i.e., weights and activations), thus reducing the  energy consumption of the DNN accelerator as well.
However, in both cases the DNN accuracy is affected and thus, methods to mitigate the associated loss are required.

Earlier works on DNN pruning attempted to maximize network sparsity or reduce MAC operations, using either a fine- or coarse-grained approach.
Though, exploiting pruning at both granularities at the same time is an uncharted territory for DNN compression.
Moreover, indirect optimization metrics (e.g., MAC count) proved suboptimal in maximizing or even achieving latency/power gains on hardware~\cite{Yang:ECCV:2018:netadapt}, mandating the need for hardware-aware compression frameworks with direct platform feedback~\cite{Yang:IEEE-CVPR:2017:energy}.
Still, combining the benefits of pruning and quantization remains mostly unexplored, mainly due to the highly increased complexity of their joint application~\cite{Wang:CVPR:2020:apq}.
To address this limitation, related works attempted to combine pruning with quantization by following a learning-based approach (e.g., with reinforcement learning (RL))~\cite{He:ECCV:2018:amc} and showed promising results.
However, in such works, fine-tuning and/or long-term retraining plays an essential role in improving upon the DNN accuracy.
Though, fine-tuning during the optimization phase and/or retraining afterwards, tremendously increases the required execution time, especially when dealing with large classification datasets (e.g., ImageNet).
In addition, such approaches may be infeasible due to proprietary or private training data~\cite{Zervakis:DAC:2021:convar}, and/or when targeting resource constrained embedded devices that do not support training.
For example, considering the latter, assume a resource constrained embedded system that comprises an inference DNN accelerator.
Deploying a hardware-aware compression optimization that requires fine-tuning/retraining would be infeasible in that scenario.
Hence, although on-device optimization would enable hardware-aware compression and usage of private local data, existing state-of-the-art techniques~\cite{He:ECCV:2018:amc, Wang:CVPR:2019:haq, Hu:AAAI:2021:opq}, due to their inherent requirement for fine-tuning, cannot be deployed on resource constrained embedded systems. 

In this work, we address the aforementioned limitations and propose a learning-based compression framework for energy-efficient DNN inference on accelerator platforms.
To achieve this, we apply layer-wise pruning and mixed-precision quantization.
Our framework explores the full spectrum of the joint pruning and quantization design space in a hardware-aware manner, without any retraining/fine-tuning (i.e., one-shot).
Our approach is generic to the pruning criterion, meaning that an optimal (different) pruning algorithm is selected per layer, from a predefined set of diverse fine- and coarse-grained pruning techniques.
To handle the huge design space, we utilize RL to automatically prune and quantize an input DNN.
We propose a novel composite RL agent, able to learn a compression policy which optimally balances the accuracy/energy trade-off.
We evaluate our framework over state-of-the-art image classification models, on the CIFAR-10, CIFAR-100, and ImageNet datasets.
Our experimental results demonstrate that the compressed DNNs generated by our framework feature small accuracy loss and up to $57$\% energy gains.

\noindent
\textbf{Our main contributions in this work are the following:}
\begin{enumerate}[wide, labelindent=0pt, label=\textbf{(\arabic*)}, ref=(\arabic*), nosep]
\item
We propose an automated framework for energy-efficient DNN inference, which applies for the first-time one-shot compression in a learning-based manner and with hardware-aware techniques.
\item We boost the pruning efficiency by considering a diverse set of pruning algorithms and combining, for the first time, fine- with coarse-grained pruning.
\item We propose a novel composite RL agent,
able to learn the optimal pruning and quantization profile for each layer.
\end{enumerate}

The rest of the paper is organized as follows:
Section~\ref{sec:related} provides an overview of the existing literature on DNN compression-related topics.
Section~\ref{sec:motivation} serves as the motivation for our work, by showcasing via examples the benefits of including diverse pruning algorithms (instead of monolithic ones) and mixed-precision quantization (instead of uniform).
In Section~\ref{sec:framework}, the core idea of our work is presented, detailing our compression technique, the implemented composite RL agent and finally, our energy estimation model.
Section~\ref{sec:evaluation} presents the evaluation setup and results, focusing on comparisons against the state of the art and the efficacy of our exploration mechanism.
Section~\ref{sec:concl} concludes this paper.
\begin{figure*}[ht]
\centering
\resizebox{0.9\textwidth}{!}{
\includegraphics{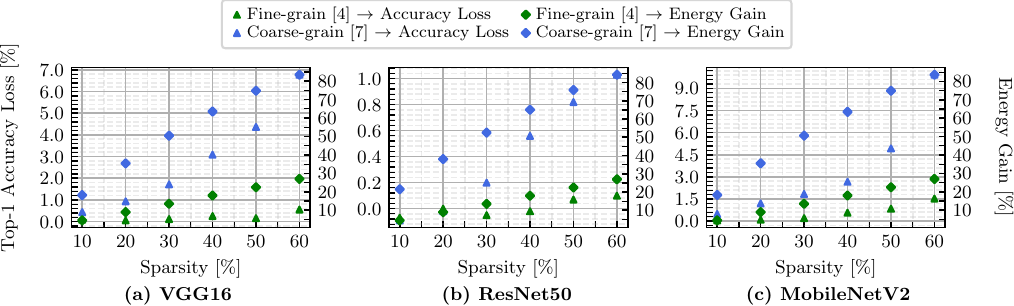}
}
\caption{Evaluation of accuracy loss (left y-axis) and energy gain (right y-axis) of different sparsity rates (x-axis) for (a) VGG16, (b) ResNet50 and (c) MobileNetV2 on CIFAR-10. Color separates fine-grain (blue) and coarse-grain pruning (green). Shape separates between accuracy loss (triangles) and energy gain (diamonds).}
\label{fig:pruning}
\end{figure*}

\section{Related Work}\label{sec:related}
Several techniques for DNN pruning and quantization have been proposed throughout the years, aiming to reduce the memory footprint while retaining acceptable prediction accuracy of DNNs.
Earlier works on pruning~\cite{Han:NIPS:2015:weights, Lee:ICLR:2019:snip} focused on fine-grained removal of connections (i.e., weight pruning) based on sensitivity analyses.
In~\cite{Frankle:ICLR:2019:lottery}, authors contemplated that randomly-initialized dense DNNs contain subnetworks, able to match the accuracy of their originator after retrained in isolation.
\cite{Yang:ECCV:2018:netadapt} presents a hardware-aware pruning framework to adapt a DNN to a given mobile platform, driven by direct hardware metrics to meet constrained budget requirements.
In~\cite{Yang:IEEE-CVPR:2017:energy}, fine-grained pruning is guided by energy estimations from arbitrary hardware platforms for energy-efficient DNN inference.
In~\cite{Niu:ASPLOS:2020:patdnn}, a methodology to prune fine-grained patterns inside coarse grain structures is proposed, which, combined with compiler-level optimization techniques, leads to inference speedup.
However, quantization is not considered by the above.

Focusing on post-training quantization, a three part methodology was proposed in~\cite{Banner:NIPS:2019:post}, composed of activation clipping, per-channel bit allocation, and weight bias correction, targeting $4$-bits precision.
However, not all layers are quantized.
Authors in~\cite{Zhao:ICLML:2019:ocs} duplicate channels containing outliers and halve their outputs, decreasing, thus, the quantization error and preserving network accuracy.
In~\cite{Gong:IEEE-CS:2021:vecq}, the quantization step and scaling factors for DNN weights are discovered through minimization of a composite vector loss instead of the quantization error.
However, the impact on hardware efficiency is not considered.

\begin{table}[!t]
\small
\setlength{\tabcolsep}{2pt}
\renewcommand{\arraystretch}{1.2}
\caption{Taxonomy of the differentiating characteristics between existing state-of-the-art works and our framework.}
\centering
\begin{tabular}{lcccc}
\hline
& AMC~\cite{He:ECCV:2018:amc} & HAQ~\cite{Wang:CVPR:2019:haq} & ASQJ~\cite{Yang:CVPR:2012:admm} & \textbf{Ours}
\\ \hline
Fine-grained Pruning & - & - & \checkmark & \cmark
\\ \hline
Coarse-grained Pruning & \checkmark & - & - & \cmark 
\\ \hline
\makecell[l]{Mixed-Precision \\ Quantization} & - & \checkmark & \checkmark & \cmark 
\\ \hline
Hardware feedback & \checkmark & \checkmark & - & \cmark
\\ \hline
Without retraining & - & - & - & \cmark
\\
\hline
\end{tabular}
\label{tab:related_comparison}
\end{table}
Improving on rule-based methods, RL has been employed to discover optimal compression solutions in a learning-based manner~\cite{He:ECCV:2018:amc, Wang:CVPR:2019:haq, Elthakeb:IEEE-Micro:2020:releq}.
Automatic model compression~\cite{He:ECCV:2018:amc} aims to provide layer-wise sparsity targets by utilizing a Deep Deterministic Policy Gradient (DDPG) agent in several hardware-emulating constraint scenarios.
In~\cite{Wang:CVPR:2019:haq}, on the other hand, an RL-based framework is proposed to find layer-wise optimal mixed-precisions, based on hardware-aware feedback from arbitrary platforms.
Both approaches consider only a single compression technique (pruning or quantization) to limit the associated design space explored by the RL agent.

Pruning and quantization are orthogonal to each other and can be applied synergistically for larger energy and memory gains.
Han et. al~\cite{Han:ICLR:2016:deep_comp} were the first to combine weight pruning with quantization (by means of clustering and weight sharing) and Huffman coding to further compress DNNs.
In~\cite{Yang:CVPR:2012:admm}, an automated framework for jointly pruning and quantization was based on alternating direction method of multipliers (ADMM).
In~\cite{Wang:CVPR:2020:apq}, subnetworks from an once-for-all network~\cite{Han:arxiv:2020:once} (found via architectural search) were sampled, quantized with mixed-precision and fed to a heuristic optimization to determine optimal compression rate and precision.
A differentiable approach is proposed in~\cite{Wang:ECCV:2020:djpq}, as a joint loss function is constructed to achieve a balance between compression techniques whilst preserving the accuracy during training.
In~\cite{Tung:IEEE-PAMI:2018:deep}, pruning and quantization are applied via clipping and quantizing the remaining weights during training, in a single stage.
Similarly, \cite{Kim:NeurIPS:2020:position} proposes a technique to scale each gradient based on the weight vector position, acting as a regularizer for pruning/quantization-aware training.
A weight representation scheme is prosed in~\cite{Kwon:CVPR:2020:structured}, tailored for irregular sparse matrices in a hardware friendly manner (i.e., exploiting XOR gates).
PQK~\cite{Kim:interspeech:2021:pqk} is a distillation-based pruning framework which utilizes (previously unimportant) pruned and quantized weights in the distillation phase, instead of pretrained teacher models.

Some related works apply pruning and quantization in one-shot, meaning that compression is applied without access to training data.
In~\cite{Hu:AAAI:2021:opq}, authors use an analytical Langrangian-based approximation model to find layer-wise pruning masks and quantization steps, followed by fine-tuning.
In~\cite{Frantar:NeurIPS:2022:obc}, an exact mathematical model is used for post-training one-shot compression, followed by a greedy error minimization algorithm.
BayesianBits~\cite{Van:NeurIPS:2020:bayesian} proposes a quantization-based compression framework, which relies on residual error addition to power-of-two weights (including $0$ for pruning).
However, most of the above techniques either involve differentiable optimization, thus moving away from the post-training setting, or require fine-tuning/retraining to recover the DNN accuracy.

\textbf{Our differentiators from the state of the art are manyfold:}
(i) we are the first to combine fine- and coarse-grained pruning techniques in our exploration phase.
As we will demonstrate, the pruning granularity choice for a given DNN is not trivial and strongly depends on the model.
(ii) We jointly prune and quantize each layer to a different sparsity ratio and precision.
(iii) We directly include hardware feedback, from a DNN accelerator during the training of the RL agent.
(iv) We do not employ any retraining or fine-tuning, during or post-exploration.
A taxonomy of the above differentiating factors is presented in Table~\ref{tab:related_comparison}.

\section{DNN Compression: Pruning and Quantization}\label{sec:motivation}
In this work, we jointly apply pruning and quantization to a given DNN with fixed architecture, aiming to reduce its energy consumption, whilst maintaining high accuracy levels.
Pruning directly compresses the model architecture, by removing parameters in a structured (i.e., coarse-grained) or sparse (i.e., fine-grained) manner.
In contrast, quantization leaves the architecture intact and only reduces the precision used for the numerical representation of its parameters.
Naturally, the two methods are non-overlapping and complement each other in reducing the resource budget of the targeted DNN.
In fact, centroid-based quantization can directly benefit from a pruned model, as the quantization error can be reduced when less centroids are used~\cite{Han:ICLR:2016:deep_comp}. 
However, both techniques exhibit a non-trivial trade-off between accuracy and energy consumption.
Thus, their combined application requires a systematic approach (more details in Section~\ref{sec:framework}).
The rest of this section explains and motivates the intuition behind employing diverse pruning techniques and mixed-precision quantization in our framework.

\subsection{Diverse Set of Pruning Techniques}\label{sec:combined}
Here, we motivate the need for employing diverse pruning algorithms, depending on the accuracy/energy trade-off of each given DNN.
Typically, pruning algorithms follow either a fine-, or a coarse-grained approach, each offering unique hardware benefits.
The intuition behind fine-grained pruning~\cite{Han:NIPS:2015:weights, Lee:ICLR:2019:snip,Guo:arxiv:2016:dynamic} derives from the observation that removing weights with small magnitude, e.g., close to zero, marginally affects the DNN accuracy.
Considering weight-stationary inference, energy savings derive from the reduced switching activity (and consequently, lower dynamic power) of the arithmetic units of the accelerator, when operating on zero inputs~\cite{Amrouch:TCAD:2021}.
Contrarily, coarse-grained pruning algorithms~\cite{Li:arxiv:2016:convnets, He:CVPR:2017:channel} remove regular structures within convolutional layers (e.g., filters and channels), thus significantly reducing the model's size and the number of operations.
In this case, the entire computation is skipped, leading to easier exploitation of pruning and direct energy gains.
Regarding compression trade-offs, weight pruning generally achieves much higher compression rate than structured pruning at comparable accuracy~\cite{Niu:ASPLOS:2020:patdnn}. 
However, fine-grained sparse models cannot be supported by off-the-shelf libraries~\cite{Ma:TNNLS:2021:non}, and specialized hardware (e.g., sparse and zero skipping accelerators) and software solutions are needed to take full advantage of the sparsity and eventually improve inference.

\begin{table}[t]
\small
\centering
\caption{The pruning algorithms considered in our work.}
\label{tab:pruning_table}
\begin{tabular}{cc}
\hline
\textbf{Algorithm} & \textbf{Pruned Patterns} \\
\hline
Sensitivity~\cite{Lee:ICLR:2019:snip} & Weights \\
Level~\cite{Han:NIPS:2015:weights} & Weights \\
Splicing~\cite{Guo:arxiv:2016:dynamic} & Weights  \\
L1-Ranked~\cite{Li:arxiv:2016:convnets} & Filters/Channels  \\
L2-Ranked~\cite{Li:arxiv:2016:convnets} & Filters/Channels  \\
Bernoulli~\cite{Pan:NPS:2020:dropfilter} & Filters  \\
FM Reconstruction~\cite{He:CVPR:2017:channel} & Channels \\
\hline
\end{tabular}
\end{table}

Our framework is build upon the state-of-the-art pruning algorithms of Table~\ref{tab:pruning_table}.
The corresponding pruned patterns of each technique are also presented.
As these techniques prune convolutional or fully-connected layers, we focus on Convolutional Neural Networks (CNNs) in our work.
Even though new pruning techniques can be seamlessly integrated, to extend the applicability of our framework to other architectures (e.g. RNNs), they are beyond the scope of this paper.
Different DNNs respond differently to each technique and a single pruning method cannot provide an optimal solution across a range of models.
Therefore, considering a diverse set of pruning techniques is crucial for the compression efficiency of our framework.

As a motivation example, Figure \ref{fig:pruning} compares the accuracy loss-energy savings trade-off for different sparsity levels (i.e., percentage of zero parameters) across varying DNNs.
The pruning algorithms ``Level''~\cite{Han:NIPS:2015:weights} (fine-grained) and ``L1-Ranked''~\cite{Li:arxiv:2016:convnets} (coarse-grained) are examined.
An Eyeriss-based~\cite{Chen:JSSC:2016:eyeriss} DNN accelerator is considered and three different DNNs, trained on the CIFAR-10 dataset, are used: (a) VGG16, (b) ResNet50, and (c) MobileNetV2.
As shown in Figure \ref{fig:pruning}, all three DNNs have different sensitivity to pruning. 
MobileNetV2 is the most sensitive one, with L1-Ranked dropping accuracy more than $1.5\%$ even for just $10\%$ sparsity. 
Contrarily, ResNet50 proves the most resilient model, offering accuracy loss of less than $1\%$ for almost all sparsity rates and both algorithms.
Overall, the coarse-grained pruning algorithm exhibits significant accuracy loss (especially for rates above $40\%$) but also higher energy savings across all examined networks.
However, this is not always the case.
Assuming a conservative accuracy loss threshold of $1\%$, ``Level''~\cite{Han:NIPS:2015:weights} for MobileNetV2 achieves $50\%$ sparsity with higher energy gain compared with the \textit{only} solution of ``L1-Ranked''~\cite{Li:arxiv:2016:convnets} (at $10\%$ sparsity) that satisfies this threshold.
Based on the aforementioned observations, the decision of a pruning technique is not trivial, it depends on the accuracy and energy requirements and is DNN-specific.
Overall, relying on a single pruning algorithm will deliver suboptimal accuracy/energy trade-offs.

\subsection{Mixed-Precision Quantization}\label{sec:quant_and_prune}
\begin{figure}[!ht]
    \centering
    \includegraphics{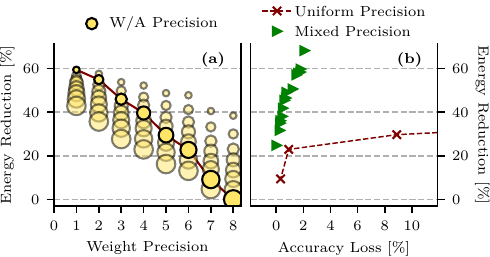}
    \caption{(a) Energy reduction due to quantization when considering a fixed precision 8-bit Eyeriss-based~\cite{Chen:JSSC:2016:eyeriss} DNN accelerator, for different weight and activation precisions. x-axis represents the precision for the weights while the radius of each circle represents the precision of the activations. The solid line represents equal weight/activation precision. (b) Energy-Accuracy trade-off when considering Uniform vs Mixed per-layer precision for ResNet18 on CIFAR-10.}
    \label{fig:quant_energy}
\end{figure}
Quantizing a (pruned) model introduces quantization errors to the (unpruned) parameters, causing a (further) detriment to the DNN accuracy.
Mixed-precision quantization can alleviate some of the associated accuracy loss (or allow for further potential gains), compared to the more common, uniform quantization.
Note, mapping the entire model to the same precision may not be optimal, since not all layers have the same sensitivity to the introduced error~\cite{Wang:CVPR:2019:haq}.

To put the energy gains of mixed-precision quantization into perspective, we assume without any loss of generality a \textit{worst-case scenario} in which the targeted DNN accelerator features a fixed architecture with single-precision MAC units, e.g.,~\cite{Chen:JSSC:2016:eyeriss,Jouppi:ISCA:2017:datacenter,Song:ISSCC:2021}.
The precision of 8 bits is selected, since it is mainstream in embedded DNN accelerators (e.g.,~\cite{Song:ISSCC:2021}), and it has been repeatedly demonstrated that 8-bit inference delivers close-to-floating-point inference accuracy~\cite{Jouppi:ISCA:2017:datacenter}.
As a result, in our worst-case scenario, computational energy gains due to mixed-precision quantization originate only from the precision-scaled inputs and the associated reduced toggling of the MAC units.
Supplying 8-bit MAC units with lower-precision inputs ($<$8bit) leads to reduced switching activity and thus lower dynamic power~\cite{Amrouch:TCAD:2021}. 
Figure~\ref{fig:quant_energy}a presents the potential energy reduction from quantizing a DNN below 8 bits, on an 8-bit DNN accelerator.
Energy gains are reported with respect to the baseline 8-bit Eyeriss-based~\cite{Chen:JSSC:2016:eyeriss} accelerator (details regarding the experimental setup and the energy evaluation are found in Section~\ref{sec:evaluation}).
As shown in Figure~\ref{fig:quant_energy}a, significant energy savings can be achieved when considering low-precision ($<$8-bit) weights and activations.
For example, for 5-bit weights and activations, $29$\% energy reduction is obtained.
Finally, Figure~\ref{fig:quant_energy}b presents an illustrative example that highlights the significance of considering mixed-precision compared to uniform quantization. 
As shown in Figure~\ref{fig:quant_energy}b, mixed-precision solutions populate a higher Pareto front, due to fine-grained manner in which the energy-accuracy space is explored.
It is worth mentioning that for only $0.5$\% accuracy loss, mixed-precision quantization achieves $38.1\%$ energy gains compared to only $9.4\%$ for uniform quantization.

Inherently, transprecision DNN accelerator architectures, e.g., \cite{RapidAI:ISCA2021,TPUv4i:ISCA2021}, can potentially exploit the benefits originating from mixed-precision quantization in full, and maximize the energy gains.
Note that although we consider a single-precision DNN accelerator, transprecision architectures can be seamlessly used in our framework due to its hardware-aware nature (see Section~\ref{sec:reward}).
In that case higher energy gains would also be expected.

\begin{figure}[t!]
\centering
\resizebox{1\columnwidth}{!}{
    \includegraphics{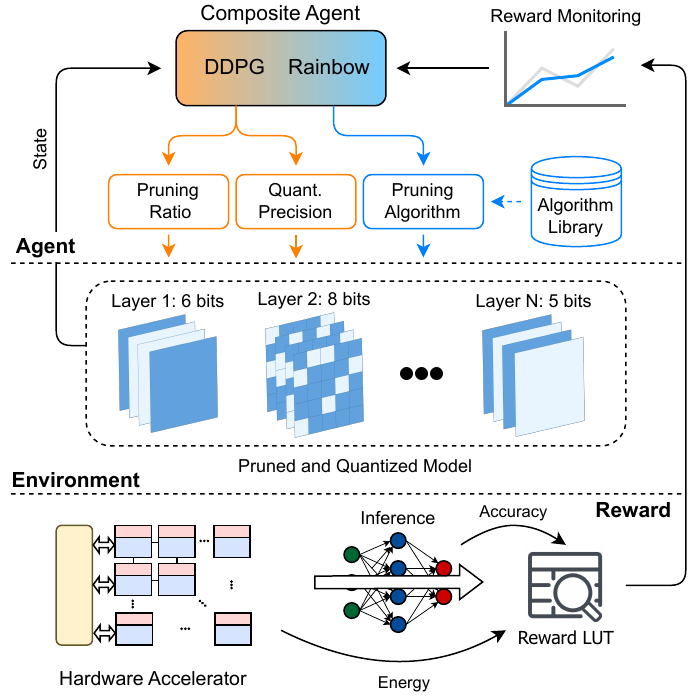}}
\caption{Abstract overview of our proposed framework.}
\label{fig:framework}
\end{figure}

\section{Proposed Hardware-Aware DNN Compression}\label{sec:framework}
In this section, the core of our proposed framework is detailed.
We tackle the hardware-aware search for a joint pruning and quantization profile of a given pretrained DNN as an RL problem.
Our goal is to learn, per DNN layer, the optimal
(a)~pruning ratio,
(b)~precision for weights and activations, and finally,
(c)~pruning technique.

An abstract overview of our framework is presented in Figure~\ref{fig:framework}.
Our composite agent receives the state of each layer and outputs three actions corresponding to its three compression directives (a-c).
Accuracy and energy consumption are measured via validation inference and a custom energy model for the targeted hardware accelerator, respectively.
Based on those measurements, we are able to deterministically formulate our reward in a LUT-based manner.
Our agent strives to maximize the accumulated reward throughout a series of training episodes.
Our framework outputs a DNN model, pruned with various techniques and quantized with mixed precision, which features a good trade-off between energy consumption and accuracy.

In contrast to the state-of-the-art approaches, our framework avoids any retraining or short-term fine-tuning.
This can potentially enable on-device optimization and DNN compression, as opposed to the computationally intensive alternative that requires fine-tuning.
For example, on-device training of the required RL-agents (see Section~\ref{sec:rl}) is significantly less computationally intensive than (re)training a DNN model (e.g., ResNet50 on ImageNet).
In addition, on-device optimization preserves data locality and therefore satisfies user privacy constraints.
Finally, the applicability of our framework is extended to scenarios where training data are either unavailable or proprietary.

The following subsections contain detailed descriptions of each aspect of our framework, as portrayed in Fig.~\ref{fig:framework}.

\subsection{Layer-wise Pruning and Quantization}\label{sec:fw_pruning}
Regarding our one-shot pruning and quantization technique, several considerations are detailed below.
Utilizing a combination of fine- and coarse-grained pruning techniques (see Table~\ref{tab:pruning_table}) comes with the additional challenge of resolving dependencies on more complex architectures (e.g. ResNets).
For example, pruning the last layer of a residual block with a structured technique enforces an identical pruning action at the shortcut layer of the same block.
Instead of simply avoiding pruning such layers, we have the flexibility to sparsify them using fine-grained techniques, thus further reducing the overall energy consumption without causing structural mismatches.
This reveals the importance of conducting layer-wise pruning (i.e., dependencies are resolved at the first dependent layer).
All layers are subject to pruning (and quantization) in our framework.

Considering that no retraining is conducted in our framework, the accuracy of the compressed DNNs will not improve post-op\-ti\-miza\-tion.
Thus, a conservative pruning policy is expected to maintain acceptable accuracy levels, especially in the coarse-grained case.
For this reason, quantization is applied upon the pruned model, as a second step, to further reduce the energy consumption.
In our framework, we apply per-channel, asymmetric, linear, post-training quantization with activation clipping based on a Laplace distribution~\cite{Banner:NIPS:2019:post}.
We set the same precision for both weights and activations, in order to reduce the vast size of our joint design space.
Note that, quantization is always applied to at least the default precision of the targeted accelerator, i.e., 8-bit in our case.

\subsection{Automated Compression with Composite RL Agent}\label{sec:rl}
To effectively traverse the large design space of possible pruning and quantization configurations, we propose a novel composite RL agent, able to learn the optimal compression profile for each DNN layer.
Our agent consists of a DDPG~\cite{Lillicrap:arxiv:2015:ddpg} and a Rainbow~\cite{Hessel:AAAI:2018:rainbow} agent, each assigned to a different task:
the DDPG algorithm is responsible for learning the pruning ratio and quantization precision, whereas Rainbow controls the pruning algorithm.
Both agent components are equipped with a prioritized replay buffer, to favor experiences with higher temporal difference (TD) error, thus improving the sampling efficiency of our framework.
A schematic overview of our RL scheme is demonstrated in Figure~\ref{fig:agent}.
In the following subsections, we describe in detail how each part of our composite agent is trained to achieve its allocated task.
\subsubsection{DDPG agent}\label{sec:ddpg}
DDPG~\cite{Lillicrap:arxiv:2015:ddpg} is an off-policy algorithm, equipped with an actor-critic model.
The actor is responsible for delivering the agent's actions (i.e., pruning ratio and quantization precision), whereas the critic is a Q-value network that determines the quality of the actor's actions, based on the gathered reward.
We enhance the learning procedure with target critic and actor networks, and the use of truncated normal distribution for noise exploration purposes.
The DDPG algorithm operates over a continuous, multidimensional action space, which in our case is a two-dimensional vector within the range of $[0,1]$ that corresponds to the pruning ratio and selected precision (i.e., the outputs of the actor).
Using a continuous space allows for finer control over both variables.
In order to translate the continuous actions to precision (i.e., bits), a simple linear mapping is required, followed by rounding to the nearest integer.
The environment state, which serves as the observation for the agent, is a $13$-dimensional vector $s_t$ and represents the full embedding of layer $t$.
In other words, the state vector mirrors every characteristic of each layer, so that the actor can act based on a rich, fully-descriptive set of features.
In the case of a convolutional layer:
\begin{align}\label{eq:state_conv}
s_t = \{t, 0, C_o, C_{in}, h_{in}, w_{in}, str, k, E_t, P_t, M_t, E^{red}_t, a_{t-1}\},
\end{align}
where $t$ is the layer index (equivalent to the agent step), $C_{in} \times h_{in} \times w_{in}$ is the size of the input feature maps (FM), $C_o$ are the output FM channels, $k$ is the kernel size, $str$ is the stride, $E_t$ is the energy consumption of (unpruned and non-quantized) layer $t$, $P_t$ is the number of its weight parameters (linked with pruning effects), $M_t$ is its memory size (i.e., $P_t \times q_b$, where $q_b=32$ precision bits for floating point weights, linked with quantization effects),
$E^{red}_t$ is the reduced energy of layer $t$ caused by actions $a_t$ and $a_{t-1}$ are the actions for layer $t-1$ (i.e., at step $t-1$).
Similarly to~\eqref{eq:state_conv}, for a fully connected layer:
\begin{align}
s_t = \{t, 1, M, N, h_{in}, w_{in}, 0, 1, E_t, P_t, E^{red}_t, a_{t-1}\},
\end{align}\label{eq:state_fc}
where $N \times M$ is the shape of the weight tensor.

\subsubsection{Rainbow agent}\label{sec:rainbow}
We implement a Rainbow agent~\cite{Hessel:AAAI:2018:rainbow} to learn the optimal pruning technique for each layer of the input DNN.
Although, both the pruning ratio and quantization precision can be easily represented as continuous variables, pruning algorithms can only be interpreted as an array of indexes.
Incrementing or decrementing indexes does not convey any physical information (importance) and is therefore hard to optimize with the continuous action space of DDPG.
For this purpose, we leverage the inherently discrete action space of the Rainbow algorithm.
Rainbow improves upon Double Q-Learning with a dueling model architecture~\cite{Wang:arxiv:2016:duel},
i.e., a value and advantage subnetworks, which introduce attention to the selected actions.
The injection of noise in the last layer of both subnetworks enriches the agent's robustness to perturbed observations from its environment.
Finally, instead of using the absolute values of the outputs from the dueling architecture, an estimation of their distribution is obtained, which boosts the agent's overall efficiency.

\begin{figure}[t!]
\centering
\resizebox{0.85\columnwidth}{!}{
    \includegraphics{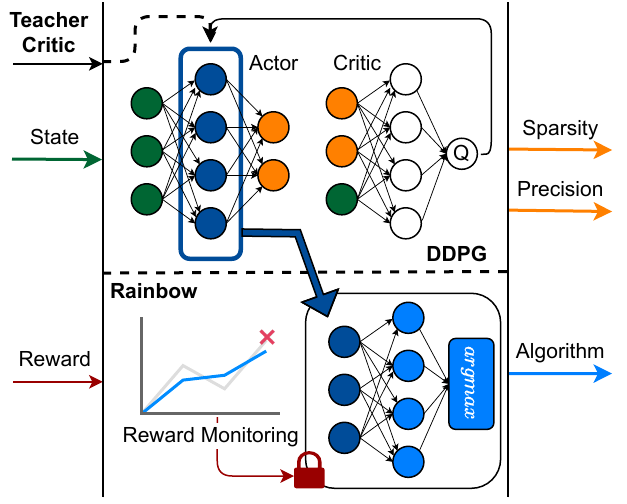}
}
\caption{
Overview of our composite RL scheme, including our two agents (DDPG and Rainbow). The reward monitoring component, which activates the Rainbow agent, and the teacher critic model which enriches the DDPG update, are also depicted.
}
\label{fig:agent}
\end{figure}

Rainbow and DDPG are directly connected, forming our composite RL agent.
The input (i.e., state vector) to the Rainbow model is the the output of the feature extractor of our compression policy, i.e. the output of the hidden layer of the DDPG actor network (see Figure~\ref{fig:agent}).
At each step, after DDPG is updated, the outputs from the hidden layer of its actor network feed the Rainbow model as inputs.
These inputs/features are used to produce the action: a discrete pruning technique.
Then, the Rainbow subnetworks (value and advantage) are allowed to update.
This feedback significantly helps the Rainbow model in learning abstract features from the DNN, without any computational overhead.

Importantly, the DDPG needs to (start to) learn the pruning sensitivity of each layer \emph{before} Rainbow appoints the optimal pruning algorithm.
Our aim is for Rainbow to safely extract patterns from its input, which would be suboptimal if the input features were rapidly changing (i.e., at the initial phase of DDPG training).
Thus, we freeze the training process of Rainbow, until the DDPG feature extractor has shown signs of improvement (i.e., increased moving average reward).
Random pruning techniques (from Table~\ref{tab:pruning_table}) are sampled at this point to remove any bias towards any specific technique.
We implement a light-weight reward monitoring scheme, which keeps Rainbow stagnant throughout the primary exploratory period of our optimization.
As soon as the reward/episode curve reflects a consistent improvement, the Rainbow agent is unlocked and allowed to control the selection of pruning techniques using the already mature features of the DDPG actor backbone.
The loss of the Rainbow agent does not back-propagate, to avoid any bias overestimation pitfalls.
Note, rewards are fed to the agent at every step, since the Rainbow model requires an update before each compression action.
Overall, the Rainbow agent learns to associate abstract features of pruning and quantization (i.e., from the last layer of the DDPG actor's feature extractor) with the best fitted technique for these features.

\subsubsection{Hardware-aware reward}\label{sec:reward}
\begin{figure}[t!]
    \centering
    \includegraphics{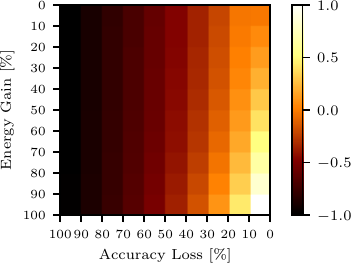}
    \caption{Heatmap of our LUT-based reward. x-axis contains the accuracy loss and y-axis the energy gain. The reward is significantly higher for solutions of accuracy loss $<\!10\%$, which is a realistic target region of our framework. For readability, the heatmap is plotted at $25\%$ of its actual resolution.}
    \label{fig:reward}
\end{figure}

We continue with the description of our reward mechanism, shared between both parts of our composite agent (i.e., DDPG and Rainbow).
We formulate our reward as a Look-up Table (LUT), indexed by accuracy and energy consumption measurements.
Accuracy is evaluated by running inference on a validation subset.
Simultaneously, we estimate the energy consumption of the pruned and quantized DNN, using a custom energy model (see Section~\ref{sec:energy_model}).
Energy measurements serve as direct feedback from the environment to the agent, as part of the reward.
Note that, although we target energy efficiency in our work, any other hardware metric (e.g., latency) is seamlessly supported since it can be measured in an identical manner.
Naively formulating our reward as an analytical notation comprising the accuracy and energy consumption (e.g., a linear combination of the two measurements), could lead to DNNs of significant accuracy loss, albeit high energy gains.
Since our framework does not retrain the final compressed model, to improve its prediction capabilities, we are realistically interested in a limited, high-accuracy region in the Pareto curve of the possible accuracy-energy trade-offs.
Thus, we create a LUT of size $40\times40$, containing a satisfactory number of accuracy loss and energy gain combinations (w.r.t. the dense model), heavily favoring, intentionally, solutions with small accuracy loss.
A heatmap of our custom reward function is illustrated in Figure~\ref{fig:reward}.
Note, a sub-sampled version of our reward function is depicted, at $25\%$ of its actual resolution, for readability purposes.
Our LUT-based reward is significantly higher when the accuracy loss is less than $10\%$.
Hence, the RL agents are strongly incentivized to prioritize actions that lead to minimal accuracy loss.
Similarly, the reward corresponding to minimal energy gains ($<\!5\%$) but also small accuracy loss ($<\!5\%$), is a small negative number, to slightly discourage RL agents from outputting close-to-zero compression actions.

\subsection{Energy Model}\label{sec:energy_model}
Here, we describe our custom energy model to estimate the energy consumption of the targeted accelerator platform and for a given pruned and quantized DNN workload.
We operate on a layer-wise fashion to extract an energy measurement for each DNN layer.
Each measurement comprises two parts: the energy consumption related to memory accesses and data movement ($E_{mem}$) and the one related to MAC computations ($E_{comp}$).
Similar to~\cite{Yang:IEEE-CVPR:2017:energy}, $E_{comp}$ is approximated as the number of MAC operations weighed by the energy cost of a single computation.
$E_{mem}$ on the other hand is calculated by extracting the total number of accesses to the accelerator's memory, multiplied by the energy cost of a single memory access.
We enhance this model by applying a multiplicative factor to both terms, which conveys the necessary information about the effects of pruning and quantization to the specific layer.
These factors ($R_x$) are appropriately named reduction coefficients, as their purpose is to apply a reduction to the energy consumption due to compression.
Thus, the total energy ($E_{total}$) of a DNN model with $L$ layers is given as follows:
\begin{align}
    E_{total} &= \sum_{l=0}^{L} E^l = \sum_{l=0}^{L} ( E^l_{mem} +  E^l_{comp} ), 
    \label{eq:e_total} \\
    E_{mem} &= \text{\#}acc * e_{mem} * R_{mem} ,
    \label{eq:e_mem} \\
    E_{comp} &= \text{\#}comp * e_{comp} *  (R_{pruned} + R_{unpruned}),
    \label{eq:e_comp}
\end{align}
where $e_{mem}$ and $e_{comp}$ are the cost values for a single memory access and MAC computation, respectively;
$R_{mem}$, $R_{pruned}$ and $R_{unpruned}$ are the reduction coefficients for the memory accesses, and for computations with pruned and unpruned parameters, respectively.
Note, the coefficients are non-unit and valued in the closed set of $[0,1]$, such that an energy reduction w.r.t the baseline is guaranteed.
Below, we analyze the calculation of each term in the equations above.

Both the number of MAC operations ($\text{\#}comp$) and memory accesses ($\text{\#}acc$) are automatically provided by an open source tool\footnote{\url{github.com/stanford-mast/nn_dataflow}}, namely NN-Dataflow.
Inspired by works related to efficient mapping of DNNs to hardware platforms~\cite{Gao:ASPLOS:2017:tetris, Gao:ASPLOS:2019:tangram}, NN-Dataflow offers a fast exploration mechanism upon diverse dataflow scheduling choices including array mapping, loop blocking and reordering, and (coarse-grained) parallel processing within and across layers.
Noticeably, the dataflow optimization techniques for intra-layer parallelism and inter-layer pipelining presented in Tangram~\cite{Gao:ASPLOS:2019:tangram} are incorporated.
Such a direct mapping and measurement helps in capturing all unique features of the accelerator, leading to an accurate modeling of energy-aware inference.
Finally, the single cost values in \eqref{eq:e_mem},\eqref{eq:e_comp} can be estimated by measuring the cost of running a single MAC (floating-point) operation on the accelerator, as well as the energy consumed by one access to memory.

Reduction coefficients incorporate the effect of pruning and quantization on each comprising term of \eqref{eq:e_total}, and their values depend on the selected pruning algorithm and precision.
Quantization is capable of reducing the computational cost of MAC operation, due to the lower precision, for all unpruned parameters.
To calculate that effect, we evaluate the energy consumption of the $8$-bit MAC units when operating over reduced-precision multiplicands (i.e., $<8$-bit weights and activations).
To that end, we design and synthesize an $8$-bit MAC circuit that consists of an $8$-bit multiplier and a $32$-bit adder to avoid accumulation overflow.
Gate-level timing simulations with realistic inputs obtained from quantized networks at different precisions are performed.
All the possible combinations for the precision of the multiplicands (i.e., $\leq8$bit) are examined, and for each combination we run a separate simulation to obtain the respective switching activity of the MAC unit.
Finally, the obtained switching activities are used to measure the power consumption of the MAC unit for each precision combination.
In this way, we are able to derive the computational power reduction ratio $R_Q$ between the baseline (i.e., $8$-bit precision for weights and activations) and any given quantization codebook (namely, $Q_W$ and $Q_A$ bits for weights and activations, respectively):
\begin{align}
    R_Q=\frac{P_{Q_W/Q_A}}{P_{8/8}}, \; R_Q \in [0, 1].
\label{eq:quant_factor}
\end{align}
Pruning effects to energy consumption differ according to the selected pruning algorithm.
Fine-grained pruning allows for moderate reduction in the computation-related energy, whilst demanding the same amount of memory accesses as the baseline, since parameters set to zero still need to be fetched from memory to calculate intermediate results for sparse matrix arithmetic operations.
Additionally, computations with pruned parameters (i.e., set to $0$) have a non-zero energy cost.
The intuition behind this statement is that when multiplying by $0$, the power consumption (and thus energy) of the MAC unit is expected to be significantly reduced but it will not be zero, due to the static power consumption as well as any toggling activity of the $32$-bit adder.
As mentioned in Section~\ref{sec:motivation}, specialized hardware (e.g., sparse and zero skipping accelerators) and software solutions are needed to take full advantage of the induced sparsity by fine-grained pruning.
To calculate this non-zero quantity, we evaluate the energy consumption of the MAC unit when one of the multiplicands is $0$ (e.g., as in fine-pruning), similarly to the aforementioned power estimation technique for quantization.
In this case, the other multiplicand and the partial sum are free to toggle.
Thus, we come up with a ``penalty'' term $P_{FG}$ which represents the average energy consumption of computations with pruned parameters compared to unpruned ones, when using a fine-grained pruning algorithm.
According to our calculations, this value is set at $0.2$, i.e., the energy used in MAC operations with weights set to $0$ is reduced by $80\%$ compared to computations with unpruned parameters.
Putting everything together, in the case of fine-grained pruning, the coefficients from~\eqref{eq:e_total},\eqref{eq:e_mem},\eqref{eq:e_comp} take the following form:
\begin{align}
\begin{split}
    R_{mem} &= 1, \\
    R_{pruned} &= P_{FG} * S, \\
    R_{unpruned} &= (1 - S) * R_Q,
\end{split}
\label{eq:rs_fg}
\end{align}
where $S$ symbolizes the fine-grained sparsity ($S \in [0, 1]$). \eqref{eq:rs_fg} shows that fine-grained algorithms may lead to moderate energy gains, due to the ``penalized'' computational energy consumed of unpruned parameters.

Coarse-grained pruning algorithms provide a straightforward reduction in both the necessary MAC computations and accesses to memory, directly proportional to the induced sparsity.
Their structured nature allows for entire blocks (i.e., filters/channels) to be removed and consequently for parts of matrix multiplication operations to be skipped.
The corresponding patterns in data movement are analogously simplified.
Summarizing, such pruning algorithms correspond to the following reduction coefficients:
\begin{align}
\begin{split}
    R_{mem} &= 1 - S, \\
    R_{pruned} &= 0, \\
    R_{unpruned} &= (1 - S) * R_Q.
\end{split}
\label{eq:rs_cg}
\end{align}
The above mathematical formulation of our energy model allows for quick energy estimations during RL training.

\section{Results and Analysis}\label{sec:evaluation}
\begin{figure}[t]
\centering
\resizebox{0.75\columnwidth}{!}{
    \includegraphics{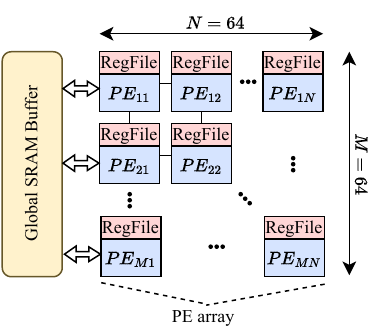}
}
\caption{Schematic overview of the Eyeriss-based\cite{Chen:JSSC:2016:eyeriss} targeted DNN accelerator for our evaluation.}
\label{fig:accelerator}
\end{figure}

\subsection{Experimental Setup}\label{sec:evalsetup}
In this section we assess the efficiency of our proposed framework by performing an in-depth evaluation over nine state-of-the-art DNNs and three image classification datasets.
For our analysis we consider:
VGG11, VGG13, and ResNet18 trained on CIFAR-10,
VGG16, ResNet34, and MobileNetV2 trained on CIFAR-100, and
VGG19, ResNet50, and SqueezeNet trained on ImageNet.
Our RL agents are trained for $1100$ episodes (first $100$ constitute the warm-up), sampling $64$ experiences per update from the prioritized replay buffer, whose size is set at $1000$ experiences.
Both actor and critic DDPG networks are built with $3$ hidden fully-connected layers of $300$ neurons and are trained with learning rates of $10^{-3}$ and $10^{-4}$, respectively.
Noise for the truncated normal distribution is initialized at $0.6$ and after the warm-up is decayed with a factor of $0.99$ after each episode.
A discount factor of $1$ is used.
The (numerous) hyperparameters required to configure the Rainbow part of our agent are directly taken from~\cite{Hessel:AAAI:2018:rainbow}.

During our optimization, rewards are computed using a $10\%$ subset of the validation set, regarding the accuracy component of the LUT.
As a target hardware platform, we consider an $8$-bit Eyeriss-based DNN accelerator~\cite{Chen:JSSC:2016:eyeriss}, which is a tile-based architecture (see Fig.~\ref{fig:accelerator}).
Specifically, we consider a $64 \times 64$ 2D array of processing elements (PEs) for each tile, with a local register file of $64$ bytes in each PE, and a shared global SRAM buffer of $32$KB, similar to~\cite{Gao:ASPLOS:2019:tangram}.
A standard 2D hierarchical memory architecture is utilized (i.e., memory channels are assumed to be on the four corners of the chip) and a total DRAM bandwidth of $3.2$Gbps.
Energy measurements from the hardware accelerator are extracted using the custom energy model described in Section~\ref{sec:energy_model}.
The MAC units are designed in Verilog-RTL, synthesized with zero-slack, and mapped to the $7$nm ASAP7 library.
The optimized arithmetic components of the commercial Synopys DesignWare library are used for the adder and multiplier. 
Synopsys Design Compiler and the \texttt{compile\_ultra} command are used for circuit synthesis, QuestaSim is used for gate-level timing simulations (and obtaining the respective switching activities), and Synopsys Primetime for measuring the power consumption.
Finally, the top-1 classification accuracy of the obtained compressed DNN is evaluated on the test set.

\begin{figure}[t]
\centering
\resizebox{0.95\columnwidth}{!}{
    \includegraphics{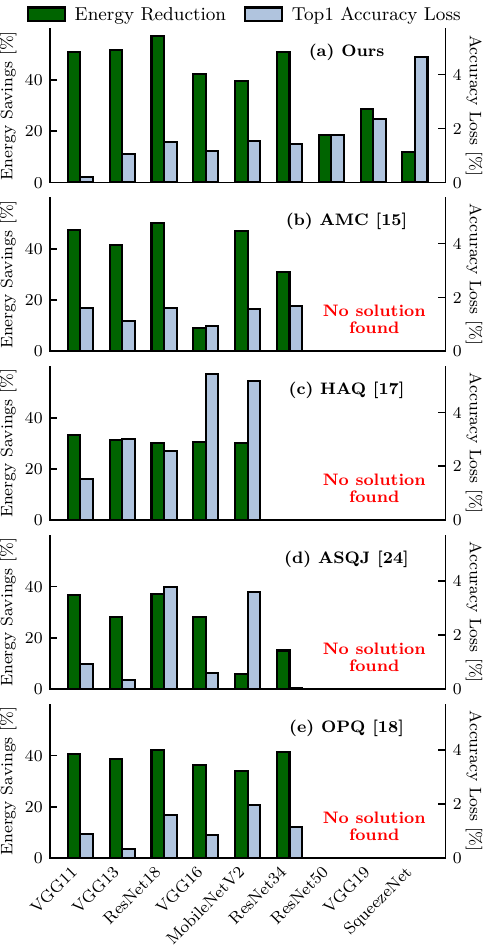}
}
\caption{
Energy gain vs accuracy loss evaluation for
(a) our framework,
(b) AMC~\cite{He:ECCV:2018:amc},
(c) HAQ~\cite{Wang:CVPR:2019:haq},
(d) ASQJ~\cite{Yang:CVPR:2012:admm} and
(e) OPQ~\cite{Hu:AAAI:2021:opq}.
Without retraining (or with only a single epoch for OPQ), the state of the art technique could not converge to a compressed DNN of acceptable accuracy loss ($<\!10\%$) for the ImageNet dataset.
}
\label{fig:sota_comparison}
\end{figure}

\subsection{Accuracy/Energy Trade-off}\label{sec:sota_comp}
To evaluate the efficiency of our framework and put the delivered energy-accuracy trade-off into perspective, we compare the DNNs produced by our framework against the following state-of-the art frameworks:
\begin{enumerate*}[label=(\roman*)]
    \item \underline{{\emph{AMC~\cite{He:ECCV:2018:amc}}}} trains a DDPG agent to apply per-layer channel pruning,
    \item \underline{{\emph{HAQ~\cite{Wang:CVPR:2019:haq}}}} applies mixed-precision quantization to weights and activations, also utilizing the DDPG algorithm,
    \item \underline{{\emph{ASQJ~\cite{Yang:CVPR:2012:admm}}}} jointly prunes and quantizes DNN weights, using the ADMM technique,
    and    
    \item \underline{\emph{OPQ~\cite{Hu:AAAI:2021:opq}}} applies one-shot pruning and quantization using an analytical model on pretrained DNN weights.
\end{enumerate*}
We select the aforementioned approaches to cover a wide spectrum of comparisons: AMC and HAQ serve as standalone approaches, applying solely pruning and quantization, respectively, in a learning-based manner.
They also constitute hardware-aware approaches (see Table~\ref{tab:related_comparison}).
Contrarily, both ASQJ and OPQ tackle the joint pruning/quantization design space, the former using ADMM and the latter with an analytical model.
Additionally, the first three approaches attempt to learn the compression profile via parameter update (i.e., RL or ADMM), whereas OPQ approximates the error minimization function via the Langrange multiplier.
Finally, OPQ is an one-shot method, i.e., no access to the training data is required to calculate the pruning masks or the quantization codebook.
All accuracy loss and energy reduction measurements below are reported w.r.t. the baseline DNN (i.e., dense DNN quantized at $8$ bits).
Since AMC~\cite{He:ECCV:2018:amc} uses floating-point inference, we quantize the resulting pruned DNN to $8$ bits.
The same quantization algorithm that is used in our framework is also used to quantize the baseline and AMC.

To the best of our knowledge our framework is the first learning-based and hardware-aware one that does not require fine-turning or retraining.
To be clear, one-shot compression via joint pruning and quantization has been proposed in prior work~\cite{Hu:AAAI:2021:opq, Frantar:NeurIPS:2022:obc, Van:NeurIPS:2020:bayesian}.
Even outside the one-shot environment, fine-tuning steps (as well as long-term retraining) are often used by related works~\cite{He:ECCV:2018:amc, Wang:CVPR:2019:haq, Yang:CVPR:2012:admm} in the exploration phase.
Retraining is orthogonal with any compression framework and acts as a standalone improvement technique to significantly recover accuracy loss without also affecting the energy efficiency of compressed model.
Thus, to ensure a fair comparison, we do not conduct long-term retraining on any compressed DNN.
Nevertheless, we allow for fine-tuning to be applied between the exploration steps of AMC~\cite{He:ECCV:2018:amc}, HAQ~\cite{Wang:CVPR:2019:haq} and ASQJ~\cite{Yang:CVPR:2012:admm}.
For OPQ~\cite{Hu:AAAI:2021:opq}, a few fine-tuning steps are used post-compression.
Although this gives a significant advantage to the state of the art, it enables us to implement the respective algorithms without tampering with their optimization flow.

Figure~\ref{fig:sota_comparison} presents our evaluation in terms of the accuracy loss vs energy gain trade-off of the compressed DNNs obtained from our framework, as well as the related works.
The derived trade-offs and comparisons are analyzed in the following subsections.

\subsubsection{Evaluation of our Framework}\label{sec:ours_eval}
As shown in Figure~\ref{fig:sota_comparison}(a), our framework delivers high energy savings and small accuracy loss across all DNNs and datasets.
A conservative threshold of $<\!2\%$ top-1 accuracy loss was imposed on our compressed DNNs for the CIFAR-10 and CIFAR-100 datasets.
This is relaxed to $<\!5\%$ for the more challenging classification task on the ImageNet dataset.
On average, our framework achieves $53.23\%$, $44.33\%$, and $19.7\%$ energy reduction for the CIFAR-10, CIFAR-100, and ImageNet DNNs, respectively.
The corresponding average top-1 accuracy loss is $0.93\%$, $1.38\%$ and $2.92\%$, on average.
As expected, the compression-induced reduction on energy consumption is at its peak for the easier classification task, on the CIFAR-10 dataset.
Error tolerance of these models stems from the fact that a substantial amount of parameters may be redundant, or quantization errors can be masked from the last layer's softmax probabilities.
Our agent exploits that fact, as the accuracy loss is incorporated in its reward, and strives to compress such models more aggressively, driving the energy gains to high levels.
On a similar but lower scale, the same can be said for the CIFAR-100 dataset.
Interestingly, the lowest obtained energy gains are observed for the MobileNetV2 architecture, which includes lightweight convolution operations (e.g., depth-wise convolution).
Such operations are constrained by design, and do not make good candidates for compressing their parameters.
We expect that a DNN architecture of moderate depth but convolution operations of high volume would be ideal for exploiting the benefits of our framework to their maximum.
A great example is the ResNet34 architecture, which even though is trained on the CIFAR-100 dataset, achieves an impressive $51.07\%$ energy gain with only $1.4\%$ top-1 accuracy loss.

Understandably, as the dataset/task becomes more complex/challenging (and since we do not use fine-tune/retraining) the obtained energy savings diminish.
Though, it is noteworthy that even for ImageNet, where energy gains are limited without retraining, our framework still manages to produce models of considerable energy savings within the strict accuracy bounds ($<5\%$).
For example, for ResNet50 on ImageNet, we achieve $18.47\%$ energy reduction and only $1.75\%$ accuracy loss.
Similarly, for VGG19 on ImageNet, the energy savings increase to $28.75$\% while the accuracy loss is $2.35$\%.
Even in the case of SqueezeNet, which is a compressed architecture by design, we achieve $11.88$\% energy gain for $4.66$\% accuracy loss.
Again, our framework delivers such energy gains by efficiently searching the joint pruning-quantization space and without requiring fine-tuning or retraining.

\subsubsection{Comparison against Standalone Approaches}\label{sec:sota_comp_standalone}
In this section, we compare our framework against the standalone pruning/quantization state-of-the-art methodologies: AMC~\cite{He:ECCV:2018:amc} and HAQ~\cite{Wang:CVPR:2019:haq}, in Figure~\ref{fig:sota_comparison}(b) and (c), respectively.
As mentioned above, both approaches involve RL-based optimization for either the per-layer pruning ration (AMC) or precision for weights and activations (HAQ), coupled with hardware-aware reward.
AMC reaches up to $49.9\%$ energy gain for ResNet18 on CIFAR-10 with $1.6\%$ accuracy loss.
On average, AMC achieves $46.32\%$ and $28.97\%$ energy reduction, respectively, for the CIFAR-10 and CIFAR-100 DNNs, while the corresponding accuracy loss is $1.44\%$ and $1.39\%$.
In comparison, our work achieves $0.51\%$ and $0.01\%$ lower top-1 loss with $6.91\%$ and $15.36\%$ higher energy gain, on average.
Quantized DNNs by HAQ feature $31.64\%$ and $30.32\%$ average energy reduction and average top-1 accuracy loss of $2.36\%$ and $5.3\%$, for CIFAR-10 and CIFAR-100.
Only a single use-case (VGG11 on CIFAR-10) adheres to the $2\%$ accuracy loss threshold set by our framework.
Interestingly, HAQ achieves semi-constant (\textapprox$30$\%) energy gains over all the examined scenarios, but with the highest accuracy loss across all techniques.
This highlights HAQ's difficulty in detecting the sensitivity of each model to quantization, and its reliance on post-optimization retraining to reduce the accuracy loss below more conservative thresholds (e.g., $1\%$).
Note, HAQ could not converge to a solution of acceptable accuracy drop ($<\!10\%$) for ResNet34 on CIFAR-100.
Similar to our framework, as the dataset complexity increases, the energy savings also decrease.

Our framework, outperforms the two related methodologies for all the examined DNNs except for AMC on MobileNetV2.
This inherently compressed architecture is amenable to structured pruning due to the regular structure of depthwise separable convolutions.
In contrast, weight pruning is considerably less impactful, as only a few redundant parameters exist.
Hence, only conservative pruning can be applied without greatly hurting prediction accuracy, which leads to reduced energy efficiency.
As a result, our framework achieves $40$\% energy reduction while AMC achieves $47$\% energy savings.
However, this difference is specific to the structure of MobileNetV2 and subject to the fine-tuning used by AMC to recover the accuracy loss, enabling AMC to further boost the applied pruning.

It must be pointed out that neither AMC nor HAQ were able to identify a solution with less than $10$\% accuracy loss for ImageNet.
Its challenging nature and the high complexity of the associated DNNs, make it significantly more difficult to apply pruning or quantization and still maintain the desired accuracy, without long-term retraining.
This further highlights the significance of all the aspects of our work, i.e., considering many pruning algorithms and intelligently selecting the optimal one, applying diverse pruning and mixed-precision quantization in a hardware-aware manner and guiding the RL exploration to high accuracy solutions via our flexible LUT-based reward.

\subsubsection{Comparison against Joint Approaches}\label{sec:sota_comp_joint}
Finally, we compare against the joint pruning/quantization state-of-the-art approaches.
Figures~\ref{fig:sota_comparison}(d) and (e) present the top-1 accuracy loss and energy gain of ASQJ~\cite{Yang:CVPR:2012:admm} and OPQ~\cite{Hu:AAAI:2021:opq}, respectively.
Note, ASQJ conducts an ADMM-based optimization where each compression stage is represented by a different variable update, whereas OPQ derives the pruning masks and quantization steps in one-shot using a Langrangian-based analytical model.
To have a fair comparison and achieve competitive results, we include $5$ retraining epochs for OPQ, after the one-shot calculations, for reasons mentioned in Section~\ref{sec:sota_comp}.
For ImageNet, this was reduced to $1$ epoch.

On average, ASQJ achieves $34.07\%$ and $16.45\%$ energy savings and top-1 accuracy loss of $ 1.66\%$ and $1.4\%$, for the CIFAR-10 and CIFAR-100 datasets, respectively.
Notice, that the energy gains obtained by ASQJ are smaller compared to both AMC and our framework.
This is expected, as fine-grained pruning (employed by ASQJ) reduces energy in a less impactful way than coarse-grained pruning (used in both our framework and AMC).
Pruned and quantized DNNs from OPQ deliver on average $40.46\%$ and $37.17\%$ with only $0.93\%$ and $1.32\%$ top-1 accuracy loss.
Such accuracy results are almost identical to the ones from our framework (see Figure~\ref{fig:sota_comparison}(a)).
Still, our energy gains for both datasets are on average $12.77$\% and $7.16\%$ higher, respectively, thus proving the effectiveness of one-shot methodologies in discovering an optimal pruning and quantization profile per layer.
Note however, OPQ requires $5$ retraining epochs to reach such accuracy levels.

Once again, the state-of-the-art did not manage to produce a DNN of acceptable accuracy levels ($<\!10\%$) on the challenging ImageNet dataset.
For ASQJ, the ADMM algorithm, despite its known capabilities, cannot traverse the design space as effectively as our RL agent, since it lacks the incentive to explore the limited region of high-accuracy solutions, which we achieve via our specialized reward function (see Section~\ref{sec:reward}).
OPQ on the other hand, even though no access to training data is needed, seems to heavily rely on fine-tuning steps, since the trend on CIFAR datasets did not continue here as well.
Contrarily, our framework operates in a truly one-shot fashion, without needing any retraining, albeit at the cost of elevated time complexity (see Section~\ref{sec:exec_time}).

\subsection{Exploration Efficacy}\label{sec:efficacy}
\subsubsection{Pruning/Quantization Analysis}
\begin{figure}[t!]
\centering
\resizebox{1\columnwidth}{!}{
    \includegraphics{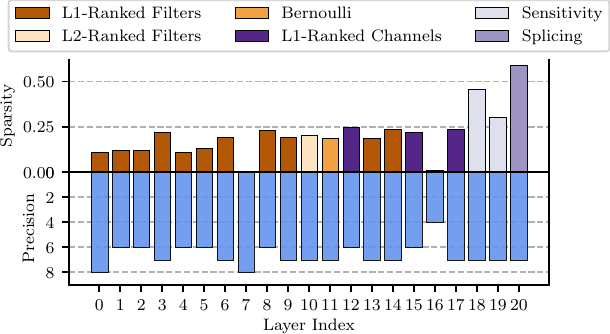}
}
\caption{Analytical demonstration of the pruning and quantization decisions made by our RL agent for the ResNet18 model on the CIFAR-10 dataset.}
\label{fig:barchart}\vspace{-2ex}
\end{figure}

In this section, we provide insight to the intuition behind our RL agent's actions, regarding the pruning ratio, quantization precision and pruning algorithm, per layer.
Figure~\ref{fig:barchart} presents a descriptive example of the solutions obtained by our framework and the derived pruning and quantization policy.
For readability reasons, ResNet18 is used (comprises only 20 layers), but similar results are obtained for the other DNNs.
Interpreting the selected pruning policy, the first layers are pruned conservatively using coarse-grained algorithms (see Table~\ref{tab:pruning_table}).
Considering the sensitivity (and therefore, lack of redundancy) of the first layers, a conservative pruning policy is a reasonable outcome, in order to retain the prediction accuracy to acceptable levels.
We observe that a series of coarse-grained algorithms (at similar pruning ratios) are selected for the middle stage of the model, since the size of each layer remains the same within that stage.
Additionally, our Rainbow agent recognizes the DDPG features corresponding to low sparsity and attributes coarse-grained pruning techniques to increase the potential energy gains.
Interestingly, the last two layers, which are fully-connected, and consequently more redundant, are pruned in a more aggressive manner with fine-grained techniques.
Our quantization policy has protected the first and last layers from aggressive quantization, keeping them at $7$ or $8$ bits.
Noticeably, the final shortcut convolutional layer (i.e., layer 16 in Figure~\ref{fig:barchart}) is barely pruned, but heavily quantized to low precision.
Extensively pruning that layer would result to accuracy degradation, due to its inter-layer dependencies to the consecutive residual block.
Thus, quantization drives the potential energy gains without significantly deteriorating the accuracy by pruning a critical layer.
The heterogeneity of sparsities/precisions directly reflects our agent's ability to identify the sensitivity in pruning and quantization of the different layers of the DNN.
Moreover, the complementary nature by which each layer is pruned and quantized stands as testament to the efficacy of our framework in intelligently exploring the joint design space.

\subsubsection{Comparison against heuristic algorithm}\label{sec:ga}
To better highlight the performance of our learning-based design space exploration, we compare our composite RL scheme against the widely adopted NSGA-II~\cite{Deb:IEEE-Evol:2002:nsga2} algorithm.
NSGA-II has prevailed as one of the most useful and fast heuristic-based exploration algorithms in the literature, due to its capability of converging quickly and generating Pareto-optimal solutions.  

NSGA-II searches for an optimal pruning/quantization configuration simultaneously for all $L$ DNN layers (contrary to the step-wise operation of RL algorithms), thus creating a genome of length $3\times L$. 
Continuous variables are therefore required, forcing the pruning algorithm variable to be controlled by rounding to discrete values which represent indices.
All (initially random) chromosomes (i.e., parent population) are subjected to the standard iterative procedure of the NSGA-II, i.e., tournament selection, simulated binary crossover and polynomial mutation.
Such operators are tailored for continuous space exploration.
From a combined pool of parent and children chromosomes, the non-dominated solutions are selected via fast non-dominated sorting and truncation based on crowding distance.
To ensure a fair comparison, the GA is allowed the same number of evaluations (i.e., energy and accuracy estimation) as our RL algorithm.
Since our agents are trained for $1100$ episodes (see Section~\ref{sec:evalsetup}), we configured the NSGA-II to run for $55$ generations, with a population size of $20$ chromosomes, to balance the exploration/exploitation trade-off.
Importantly, all the evaluations are conducted in the same manner as our technique, including our hardware aware reward, which serves as the fitness function\footnote{Since NSGA-II strives to minimize the fitness objectives, we use the inverse reward (i.e., multiplied by $-1$).} (i.e., a single fitness objective is conducted).
The output of the genetic algorithm is the single solution (pruned/quantized DNN) with the highest reward value (see Section~\ref{sec:reward}).

\begin{figure}[t!]
\centering
\resizebox{1\columnwidth}{!}{
    \includegraphics{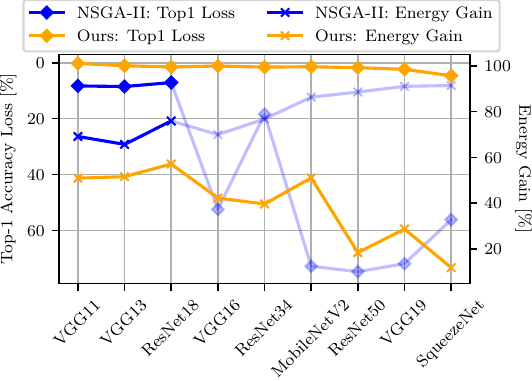}
}
\caption{
Comparative evaluation between our composite RL agent and the NSGA-II algorithm~\cite{Deb:IEEE-Evol:2002:nsga2}, in terms of top-1 accuracy loss and energy gain.
NSGA-II does not manage to produce DNNs of acceptable accuracy loss ($<\!5\%$), and the CIFAR-10 networks are the only ones with loss lower than $10\%$.
Faded lines and markers represent DNNs of unacceptable accuracy loss ($>10\%$).
}
\label{fig:ga}
\end{figure}

We present the comparative evaluation results in Figure~\ref{fig:ga}.
Overall, NSGA-II produces DNNs of high accuracy loss, albeit elevated energy gains.
For CIFAR-10, which is a more error-tolerant task (classification to 10 classes), a significant (but not destructive) top-1 loss of \textapprox$8\%$ on average is observed.
For the rest of the extracted models, for the more demanding datasets (with faded markers in Figure~\ref{fig:ga}), the best NSGA-II solution fails to limit the classification accuracy below $10\%$.
This can be attributed to the limited amount of allowed evaluations and the imposed tight accuracy constraints.
As mentioned in Section~\ref{sec:reward}, the area of interest for our reward is a very limited, high-accuracy region in the Pareto curve of the possible accuracy-energy trade-offs.
That is because training data are not used and therefore, the possibility of improving the final accuracy after pruning or quantization is absent.
Our learning-based exploration is able to guide the RL agent towards (useful) solutions of high accuracy (which represent a very small sample of our entire design space),  even at the cost of moderate energy gains (e.g., for the ImageNet dataset). 
Genetic algorithms are less sample efficient (since they heavily rely on stochastic processes) in favor of quick execution, and cannot respond well when having to deal with a limited amount of evaluations.

\subsubsection{Execution Time and Memory Requirements}\label{sec:exec_time}
Finally, we evaluate the execution time and memory requirements of our framework, and specifically, our RL-based optimization, in comparison to state-of-the-art techniques~\cite{He:ECCV:2018:amc,Wang:CVPR:2019:haq,Yang:CVPR:2012:admm, Hu:AAAI:2021:opq}.
Since the number of iterations (e.g., RL training episodes) is a user-defined parameter, our comparisons involve a single iteration for each studied method.
To ensure a fair comparison against the state of the art, the same contributing factors to the performance of each technique were used (e.g., number of worker threads, batch size, etc.).
Measurements were conducted on a desktop computer featuring an NVIDIA GeForce RTX $2080$ Super GPU operating at $1.65$ GHz and $32$GB of RAM.
Table~\ref{tab:execution_time} presents the normalized (w.r.t. the fastest technique) average execution time of a single iteration for all evaluated models on each dataset, taken from an average over $50$ iterations, to avoid statistical bias.
Table~\ref{tab:mem_req} presents a comparison of the memory utilization for the studied techniques.
\begin{table}[!t]
\small
\setlength{\tabcolsep}{2.1pt}
\renewcommand{\arraystretch}{1.2}
\caption{Comparative evaluation for the execution time of a single iteration among our proposed framework and state-of-the-art techniques. The average execution time over all evaluated models on each dataset is obtained. The fastest technique is used as the normalization baseline for each dataset.
}
\centering
\begin{tabular}{lccccc}
\hline
\textbf{Dataset} & \textbf{Ours} & \textbf{AMC~\cite{He:ECCV:2018:amc}} & \textbf{HAQ~\cite{Wang:CVPR:2019:haq}} & \textbf{ASQJ~\cite{Yang:CVPR:2012:admm}} & \textbf{OPQ~\cite{Hu:AAAI:2021:opq}}
\\ \hline
CIFAR-10 & $5.61$x & $5.60$x & $2.743$x & $19.89$x & $\bm{1.00}$\textbf{x}
\\ \hline
CIFAR-100 & $29.35$x & $12.85$x & $4.94$x & $19.82$x & $\bm{1.00}$\textbf{x}
\\ \hline
ImageNet & $5.67$x & $1.39$x & $2.82$x & $39.62$x & $\bm{1.00}$\textbf{x}
\\ \hline
\end{tabular}
\label{tab:execution_time}
\end{table}
\begin{table}[h]
\small
\setlength{\tabcolsep}{2.5pt}
\renewcommand{\arraystretch}{1.2}
\caption{
Memory requirements for the execution of a single iteration of our proposed framework and state-of-the-art techniques.
The average memory utilization per dataset is reported (using Python's memory profiler package), normalized w.r.t the lowest utilization among all methods.
}
\centering
\begin{tabular}{lccccc}
\hline
\textbf{Dataset} & \textbf{Ours} & 
\textbf{AMC~\cite{He:ECCV:2018:amc}} & \textbf{HAQ~\cite{Wang:CVPR:2019:haq}} & \textbf{ASQJ~\cite{Yang:CVPR:2012:admm}} & \textbf{OPQ~\cite{Hu:AAAI:2021:opq}}
\\ \hline
CIFAR-10 & $1.19$x & $1.08$x & $1.34$x &  $\bm{1.00}$\textbf{x} & $1.15$x
\\ \hline
CIFAR-100 & $1.20$x & $1.09$x & $1.34$x &  $\bm{1.00}$\textbf{x} & $1.15$x
\\ \hline
ImageNet & $1.31$x &  $\bm{1.00}$\textbf{x} & $1.50$x & $1.15$x & $1.74$x
\\ \hline
\end{tabular}
\label{tab:mem_req}
\end{table}

Overall, the reported execution time of our framework is on the higher end compared to the state of the art, whilst having comparable memory requirements.
OPQ~\cite{Hu:AAAI:2021:opq} presents the lowest time complexity due to its quick one-shot mathematical analysis of pruning/quantization, which translates though to elevated memory utilization for ImageNet, mostly due to multiple copies of weight tensors.
ASQJ's~\cite{Yang:CVPR:2012:admm} memory efficiency due to the sequential nature of parameter update of the ADMM algorithm proves to be time consuming.
Our framework's overhead is governed by the RL agent's update scheme, and the reward calculation (mainly its accuracy term), which has to be repeated at each step, for the reasons outlined in Section~\ref{sec:rainbow}.
The latter leads to elevated time complexity, which is evident for the complex and deep MobileNetV2 architecture.
Such calculations are also present in the state-of-the-art approaches, which helps keep the memory utilization measurements at similar levels.
So, even though we do not possess an advantage in these comparisons, the overhead introduced by our technique is on par with the state of the art.
Note, exploring the joint pruning and quantization design space gives an immediate disadvantage to our proposed technique compared to standalone approaches~\cite{He:ECCV:2018:amc, Wang:CVPR:2019:haq}.
Importantly, most related approaches require post-optimization fine-tuning, which significantly increases their memory utilization, especially with the high level of data parallelism normally used for modern datasets.
Overhead reduction, as well as improving the scalability of our technique is left for future work.

\section{Conclusion}\label{sec:concl}
In this work, we propose an automated hardware-aware framework to holistically compress DNNs for energy-efficient inference.
We explore the joint design space of fine- and coarse-grained pruning and mixed-precision quantization using a novel, composite RL-agent with a custom LUT-based reward.
Our framework does not require any fine-tuning and/or long-term retraining steps thus satisfying proprietary and/or privacy constraints.
Targeting to reduce energy consumption whilst minimally affecting prediction accuracy, we exceed the state-of-the-art compression techniques (that use fine-tuning).
Our framework achieves \textapprox$39$\% average energy reduction across all studied DNNs.


\begin{IEEEbiography}[{\includegraphics[width=1in,height=1.25in,clip,keepaspectratio]{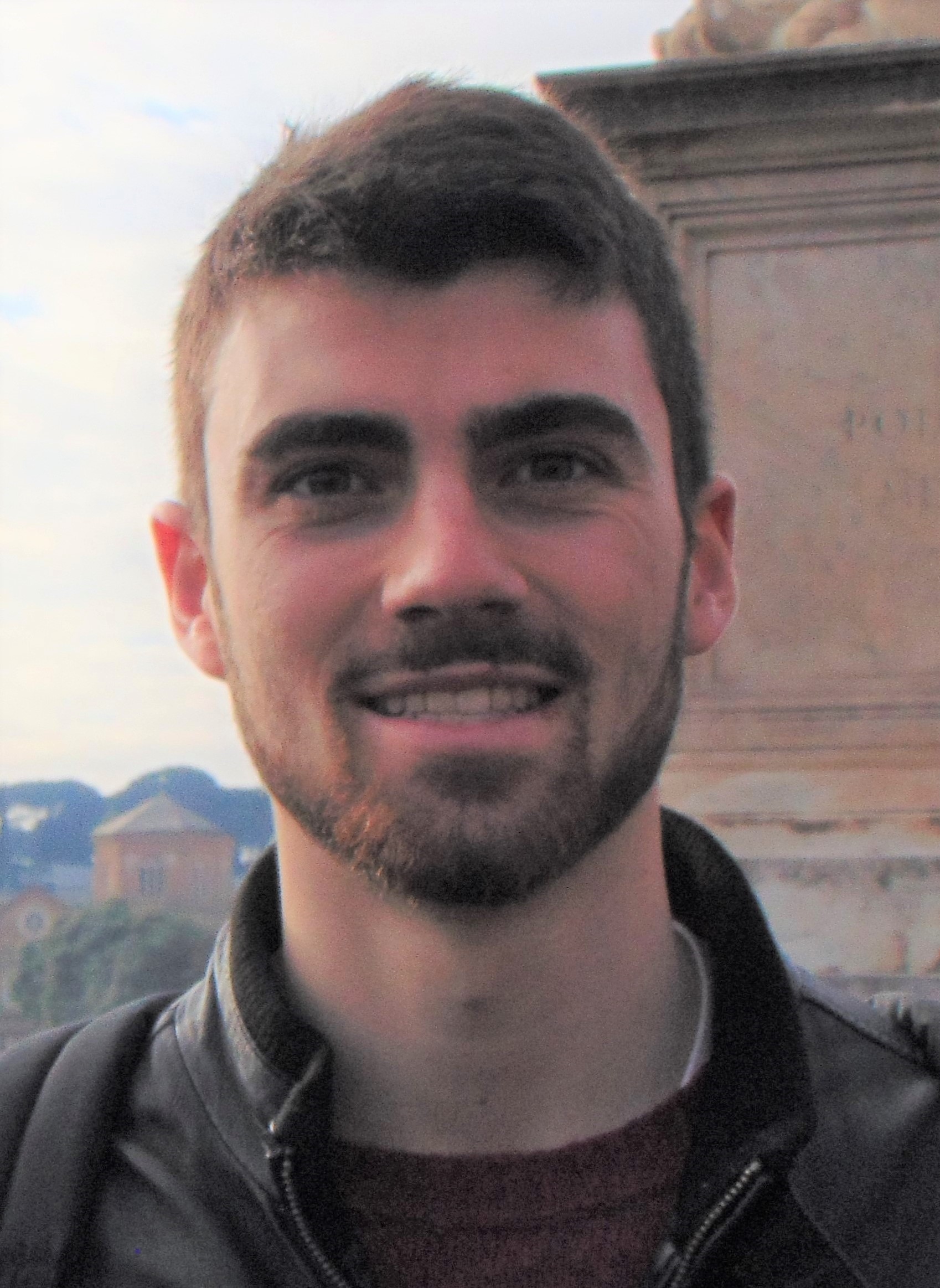}}]{Konstantinos Balaskas} received his Bachelor Degree in Physics and Master Degree in Electronic Physics from the Aristotle University of Thessaloniki in 2018 and 2020, respectively.
Currently, he is a pursuing the PhD degree at the same institution, jointly with the Chair for Embedded Systems (CES) at the Karlsruhe Institute of Technology (KIT), Germany. 
His research interests include machine learning for resource-constrained platforms and physical-driven approximate computing.
\end{IEEEbiography}

\begin{IEEEbiography}[{\includegraphics[width=1in,height=1.25in,clip,keepaspectratio]{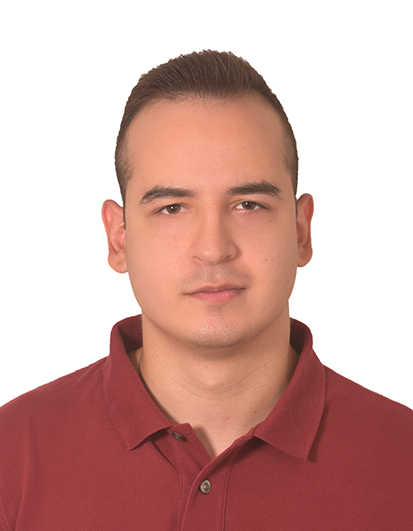}}]{Andreas Karatzas} received the Integrated Master degree (Diploma) from the department of Computer Engineering and Informatics (CEID), University of Patras, Patras, Greece, in 2021. He is currently pursuing the Ph.D. degree at the School of Electrical, Computer and Biomedical Engineering at Southern Illinois University, Carbondale, Illinois, as a member of the Embedded Systems Software Lab. His research interests include embedded systems, approximate computing, and deep learning. 
\end{IEEEbiography}

\begin{IEEEbiography}[{\includegraphics[width=1in,height=1.25in,clip,keepaspectratio]{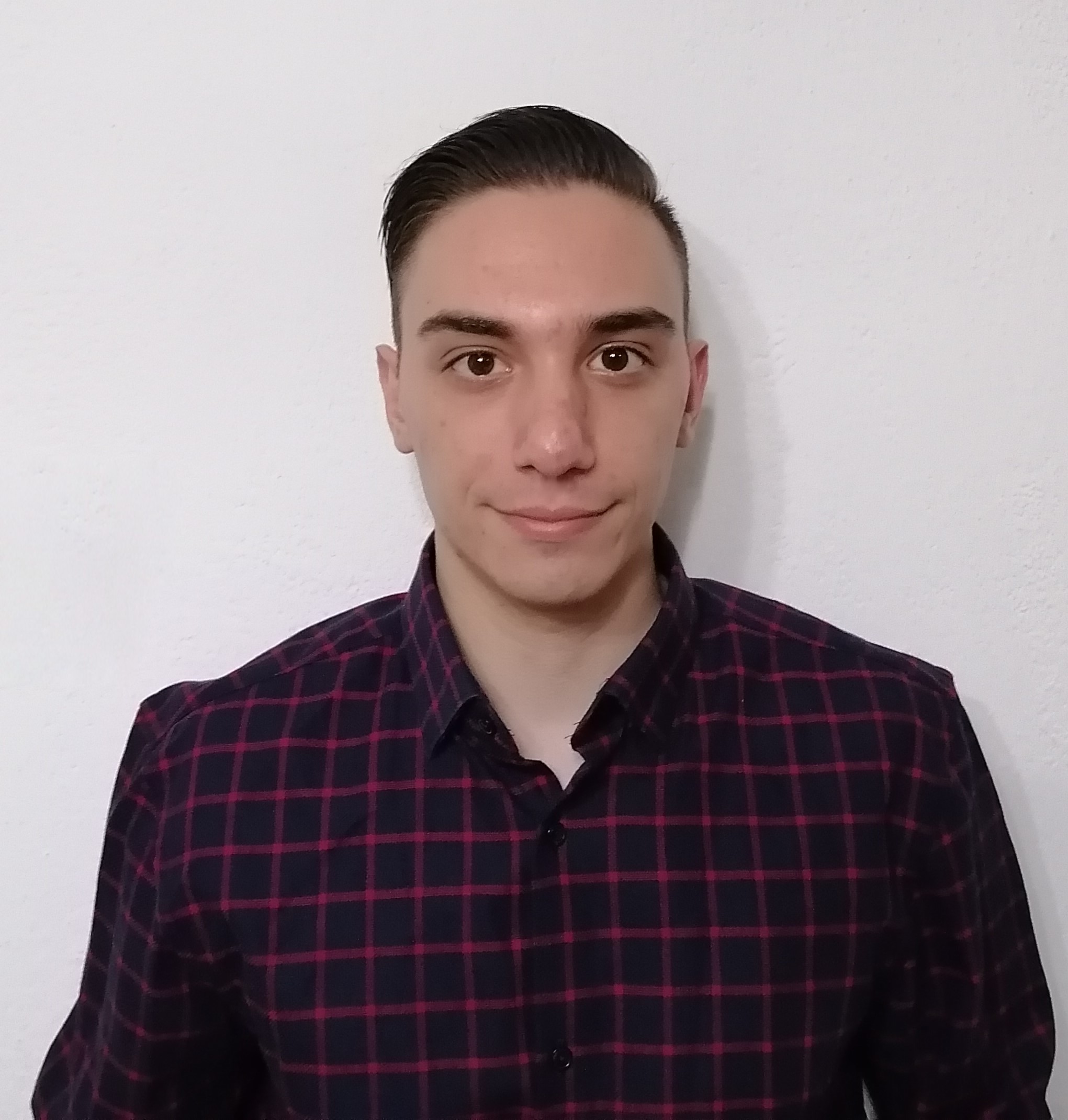}}]{Christos Sad} received the Integrated Master degree (Diploma) from the department of Electrical and Computer Engineering (ECE), Aristotle University of Thessaloniki(AUTh), in 2021. He is currently pursuing the Ph.D. degree at the School of Physics, department of Electronics and Computers, Aristotle University of Thessaloniki(AUTh). His research interests include optimization, approximate computing, neural architecture search, GPU programming and GPU acceleration.
\end{IEEEbiography}

\begin{IEEEbiography}[{\includegraphics[width=1in,height=1.25in,clip,keepaspectratio]{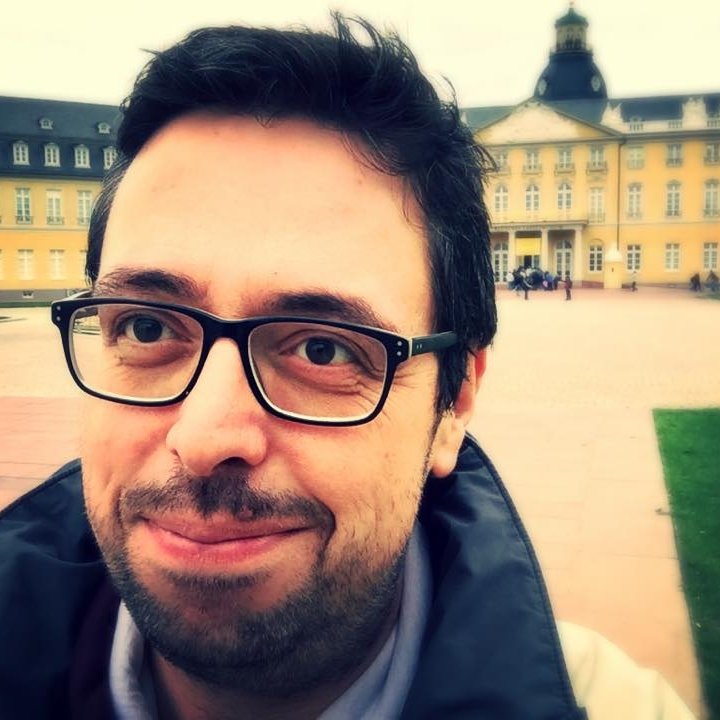}}]{Kostas Siozios} 
received his Diploma, Master and Ph.D. Degree in Electrical and Computer Engineering from the Democritus University of Thrace, Greece, in 2001, 2003 and 2009, respectively. Currently, he is Associate Professor at Department of Physics, Aristotle University of Thessaloniki. His research interests include Digital Design, Hardware Accelerators, Resource Allocation and Machine Learning. Dr. Siozios has published more than 170 papers in peer-reviewed journals and conferences. Also, he has contributed in 5 books of Kluwer and Springer. He has worked as Project Coordinator, Technical Manager or Principal Investigator in 28 research projects funded from the European Commission (EC), European Space Agency (ESA), as well as National Funding.
\end{IEEEbiography}

\begin{IEEEbiography}[{\includegraphics[width=1in,height=1.25in,clip,keepaspectratio]{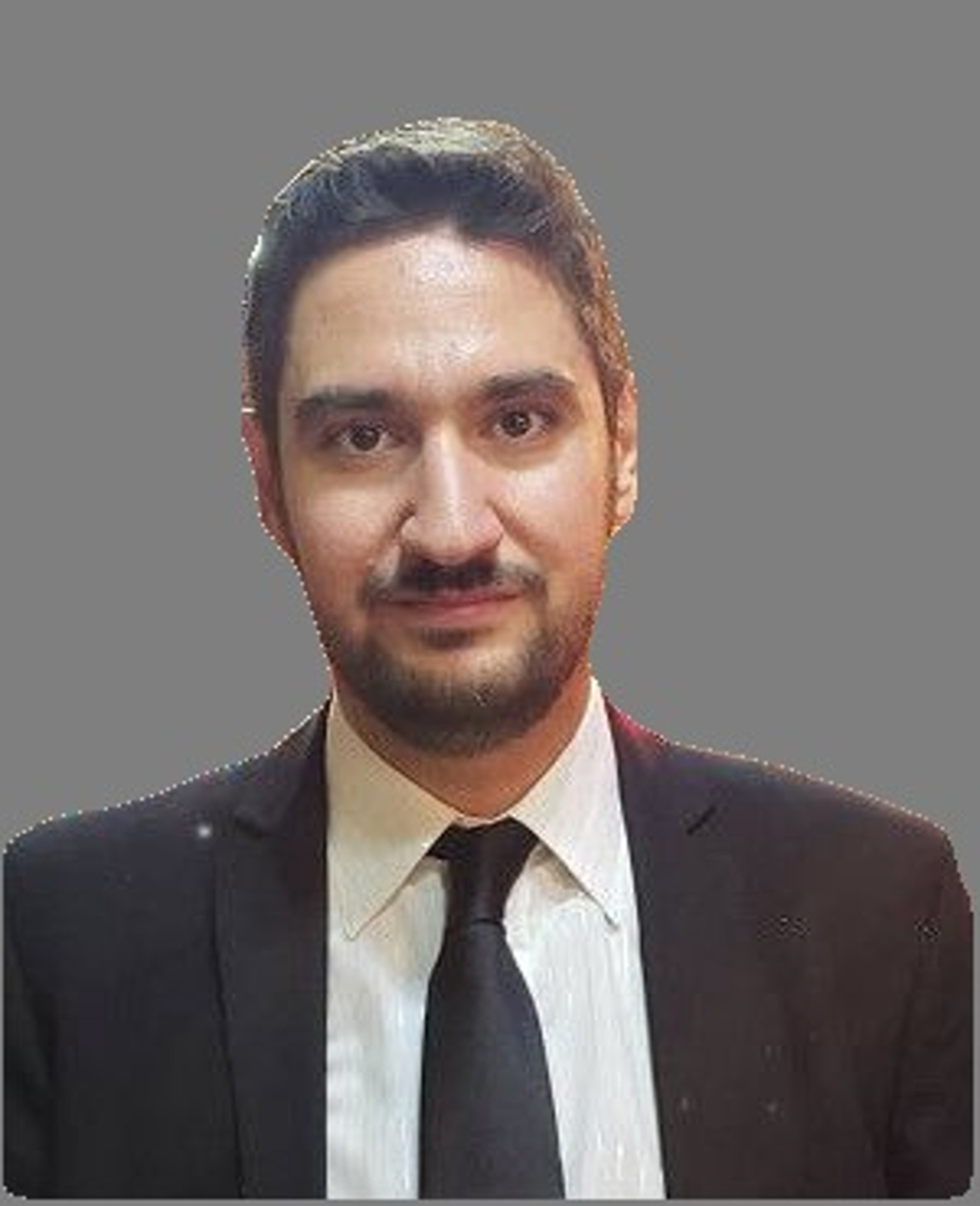}}]{Iraklis Anagnostopoulos}
is an Associate Professor at the School of Electrical, Computer and Biomedical Engineering at Southern Illinois University, Carbondale. 
He is the director of the Embedded Systems Software Lab, which works on run-time resource management of modern and heterogeneous embedded many-core architectures, and he is also affiliated with the Center for Embedded Systems. 
He received his Ph.D. in the Microprocessors and Digital Systems Laboratory of National Technical University of Athens. 
His research interests lie in the area of approximate computing, heterogeneous hardware accelerators, and hardware/software co-design.
\end{IEEEbiography}

\begin{IEEEbiography}[{\includegraphics[width=1in,height=1.25in,clip,keepaspectratio]{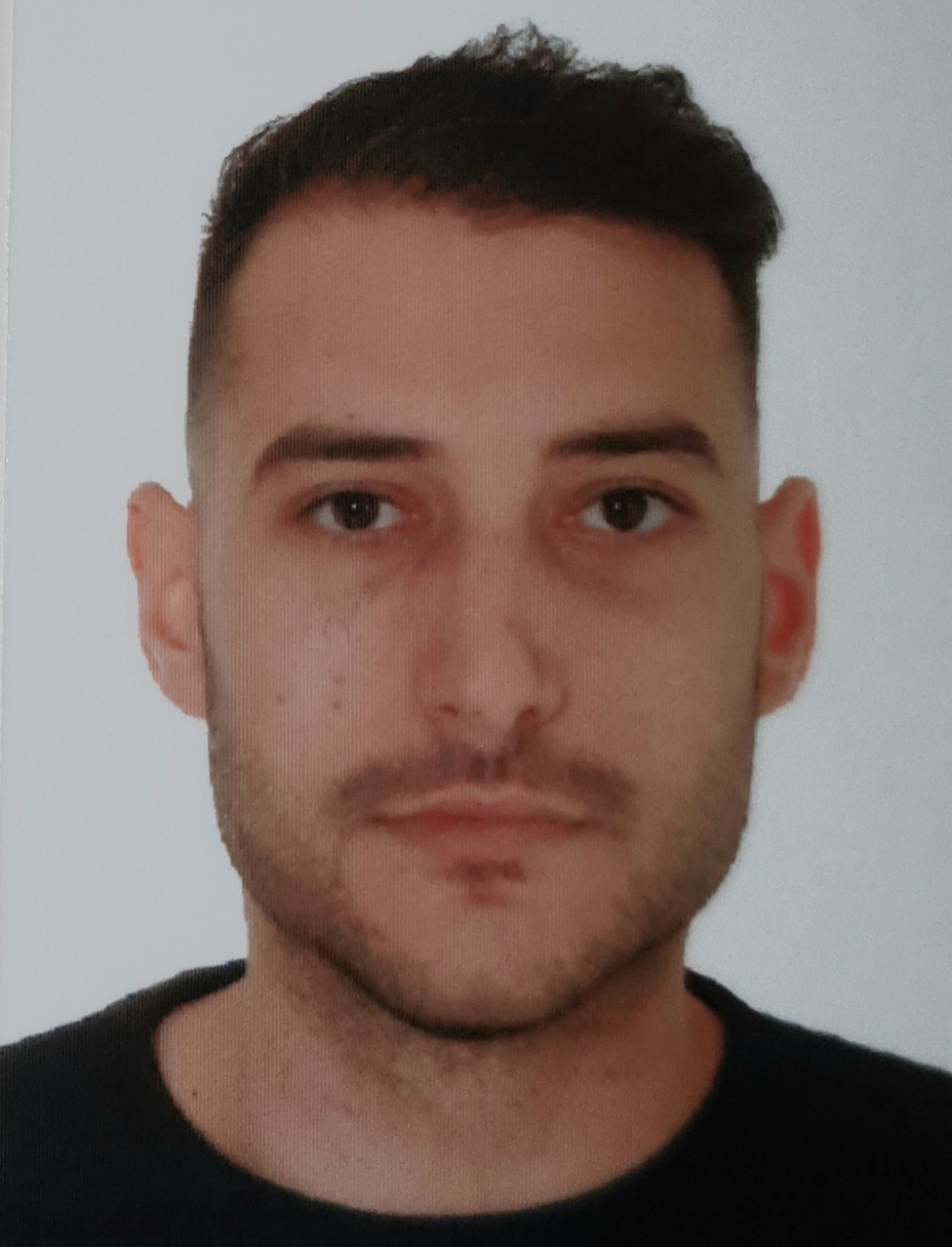}}] {Georgios Zervakis} is an Assistant Professor at the University of Patras.
Before that he was a Research Group Leader at the Chair for Embedded Systems (CES), at the Karlsruhe Institute of Technology (KIT) from 2019 to 2022.
He received the Diploma and Ph.D. degrees from the School of Electrical and Computer Engineering (ECE), National Technical University of Athens (NTUA), Greece, in 2012 and 2018, respectively.
Dr. Zervakis serves as a reviewer in many IEEE and ACM journals and is also a member of the technical program committee of several major design conferences.
He has received one best paper nomination at DATE 2022.
His main research interests include low-power design, accelerator microarchitectures, approximate computing, and machine learning.
\end{IEEEbiography}

\begin{IEEEbiography}[{\includegraphics[width=1in,height=1.25in,clip,keepaspectratio]{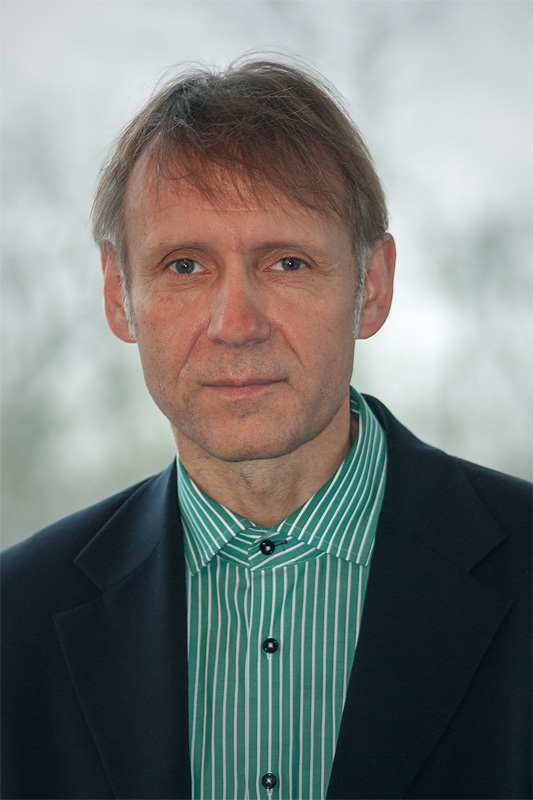}}]{J\"org Henkel} (M'95-SM'01-F'15) is with Karlsruhe Institute of Technology and was before a research staff member at NEC Laboratories, Princeton, NJ.
He has received six best paper awards from, among others, ICCAD, ESWeek and DATE.
For two terms he served as the Editor-in-Chief for the ACM Transactions on Embedded Computing Systems.
He is currently the Editor-in-Chief of the IEEE Design\&Test Magazine and is/has been Associate Editor for major ACM and IEEE Journals.
He has led several conferences as a General Chair incl. ICCAD, ESWeek and serves as Steering Committee chair/member for leading conferences and journals for embedded and cyber-physical systems. Prof. Henkel coordinates the DFG program SPP 1500 ``Dependable Embedded Systems'' and is a site coordinator of the DFG TR89 collaborative research center ``Invasive Computing''. He is the chairman of the IEEE Computer Society, Germany Chapter, and a Fellow of the IEEE.
\end{IEEEbiography}

\end{document}